\documentclass{article}

\usepackage[final,nonatbib]{nips_2016}

\usepackage{bm,amsmath,amssymb}
\usepackage{graphicx}
\usepackage[bookmarks=false]{hyperref}
\usepackage{longtable,lscape}
\usepackage[normalem]{ulem}
\usepackage{booktabs}
\usepackage{xspace}
\usepackage{xcolor}
\definecolor{mydarkblue}{rgb}{0,0.08,0.45}
\hypersetup{ 
    colorlinks,
    citecolor=mydarkblue,
    filecolor=mydarkblue,
    linkcolor=mydarkblue,
    urlcolor=mydarkblue 
}
\usepackage{subcaption}
\usepackage{url}

\usepackage[utf8]{inputenc} 
\usepackage[T1]{fontenc}    
\usepackage{booktabs}       
\usepackage{amsfonts}       
\usepackage{nicefrac}       
\usepackage{microtype}      

\usepackage{bm}
\usepackage{enumitem}
\usepackage{caption}

\newcommand{\Wb}{\mathbf{W}}

\newcommand{\bb}{\mathbf{b}}
\newcommand{\cbb}{\mathbf{c}}

\newcommand{\gb}{\mathbf{g}}
\newcommand{\hb}{\mathbf{h}}

\newcommand{\vb}{\mathbf{v}}

\newcommand{\wb}{\mathbf{w}}
\newcommand{\xb}{\mathbf{x}}
\newcommand{\yb}{\mathbf{y}}

\newcommand{\soft}{\mathrm{Softmax}}

\newcommand{\alg}{\texttt{RETAIN}\xspace}

\title{\alg: An Interpretable Predictive Model for Healthcare using Reverse Time Attention Mechanism}

\author{
Edward Choi$^*$, Mohammad Taha Bahadori$^*$, Joshua A. Kulas$^*$, \\
\normalsize \textbf{Andy Schuetz$^\dagger$, Walter F. Stewart$^\dagger$, Jimeng Sun$^*$}\\
$^*$ Georgia Institute of Technology \qquad $^\dagger$ Sutter Health \\ 
\small \texttt{\{mp2893,bahadori,jkulas3\}@gatech.edu},\\ 
\small \texttt{\{schueta1,stewarwf\}@sutterhealth.org}, \texttt{jsun@cc.gatech.edu} \normalsize
}

\begin{document}

\maketitle

\vspace*{-3mm}
\begin{abstract}
\vspace*{-3mm}
Accuracy and interpretability are two dominant features of successful predictive models. Typically, a choice must be made in favor of complex black box models such as recurrent neural networks (RNN) for accuracy versus less accurate but more interpretable traditional models such as logistic regression. This tradeoff poses challenges in medicine where both accuracy and interpretability are important.
We addressed this challenge by developing the REverse Time AttentIoN model (\alg) for application to Electronic Health Records (EHR) data. \alg achieves high accuracy while remaining clinically interpretable and is based on a two-level neural attention model that detects influential past visits and significant clinical variables within those visits (e.g. key diagnoses). \alg mimics physician practice by attending the EHR data in a reverse time order so that recent clinical visits are likely to receive higher attention. \alg was tested on a large health system EHR dataset with 14 million visits completed by 263K patients over an 8 year period and demonstrated predictive accuracy and computational scalability comparable to state-of-the-art methods such as RNN, and ease of interpretability comparable to traditional models.
\end{abstract}

\section{Introduction}
\label{sec:intro}

The broad adoption of Electronic Health Record (EHR) systems has opened the possibility of applying clinical predictive models to improve the quality of clinical care.
Several systematic reviews have underlined the care quality improvement using predictive analysis \cite{chaudhry2006systematic,jha2009use,black2011impact,goldzweig2009costs}. EHR data can be represented as temporal sequences of high-dimensional clinical variables (e.g., diagnoses, medications and procedures), where the sequence ensemble represents  the documented content of medical visits from a single patient. Traditional machine learning tools summarize this ensemble into aggregate features, ignoring the temporal and sequence relationships among the feature elements. The opportunity to improve both predictive accuracy and interpretability is likely to derive from effectively modeling temporality and high-dimensionality of these event sequences.

Accuracy and interpretability are two dominant features of  successful predictive models. There is a common belief that one has to trade accuracy for interpretability using one of three types of  traditional  models~\cite{caruana2015intelligible}:
1) identifying a set of rules (\textit{e.g.} via  decision trees \cite{kho_use_2012}), 2) case-based reasoning by finding similar patients (\textit{e.g.} via $k$-nearest neighbors \cite{gallego_bringing_2015} and distance metric learning \cite{sun2012supervised}), and 3) identifying a list of risk factors (\textit{e.g.} via LASSO coefficients \cite{fleisher_clinical_2007}). 
While interpretable, all of these models rely on aggregated features, ignoring the temporal relation among features inherent to EHR data. As a consequence, model accuracy is sub-optimal. Latent-variable time-series models, such as \cite{saria2010learning,schulam2015probabilistic}, account for temporality, but often have limited interpretation due to abstract state variables. 

Recently, recurrent neural networks (RNN) have been successfully applied in modeling sequential EHR data to predict diagnoses \cite{lipton2015learning} and model encounter sequences \cite{choi2015doctor, esteban2016predicting}. But, the gain in accuracy from use of RNNs is at the cost of model output that is notoriously difficult to interpret. While there have been several attempts at directly interpreting RNNs \cite{ghosh1992sequence,karpathy2015visualizing,che2015distilling}, these methods are not sufficiently developed for  application in clinical care.


\begin{figure}
    \centering
    \begin{subfigure}[b]{0.25\textwidth}
        \includegraphics[scale=0.35]{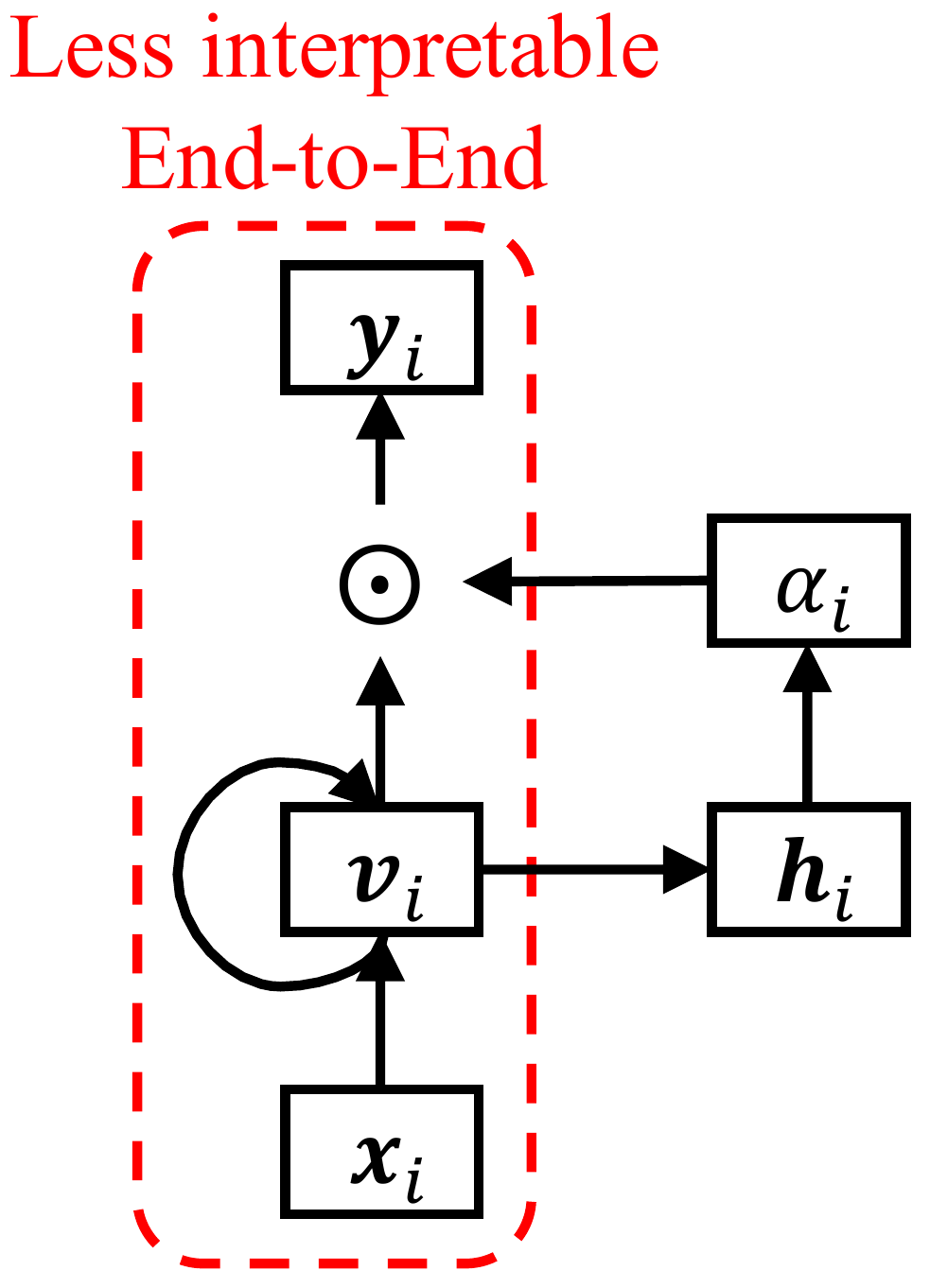}
        \caption{Standard attention model}
        \label{fig:common}
    \end{subfigure}
    \qquad \qquad 
    \begin{subfigure}[b]{0.25\textwidth}
        \includegraphics[scale=0.35]{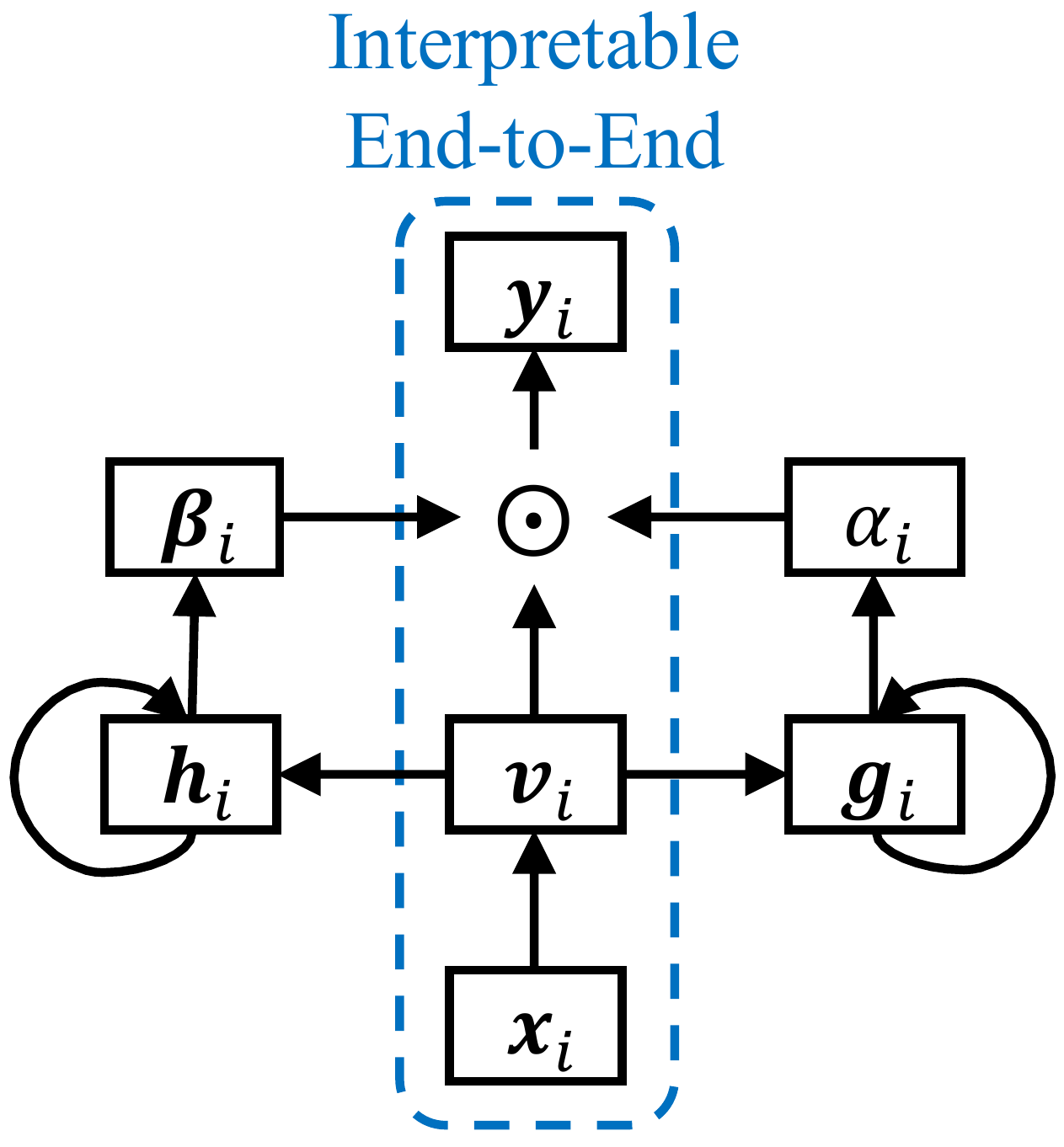}
        \caption{\alg model}
        \label{fig:retain}
    \end{subfigure}
    \caption{ Common attention models vs. \alg, using folded diagrams of RNNs.  (\subref{fig:common}) Standard attention mechanism: the recurrence on the hidden state vector $\vb_i$ hinders interpretation of the model. (\subref{fig:retain}) Attention mechanism in \alg: The recurrence is on the attention generation components ($\hb_i$ or $\gb_i$) while the hidden state $\vb_i$ is generated by a simpler more interpretable output. 
    }
    \label{fig:idea}
\end{figure}

We have addressed this limitation using a modeling strategy known as \alg, a two-level neural attention model for sequential data that provides detailed interpretation of the prediction results while retaining the prediction accuracy comparable to RNN. To this end, \alg relies on an attention mechanism modeled to represent the behavior of physicians during an encounter. 
A distinguishing feature of \alg (see Figure \ref{fig:idea}) is to leverage sequence information using an attention generation mechanism, while learning an interpretable representation. And emulating physician behaviors, \alg examines a patient’s past visits in reverse time order, facilitating a more stable attention generation. As a result, \alg identifies the most meaningful visits and quantifies visit specific features that contribute to the prediction.

\alg was tested on a large health system EHR dataset with 14 million visits completed by 263K patients over an 8 year period. We compared predictive accuracy of \alg to traditional machine learning methods and to RNN variants using a case-control dataset to predict a future diagnosis of heart failure. The comparative analysis demonstrates that \alg achieves comparable performance to RNN in both accuracy and speed and significantly outperforms traditional models. Moreover, using a concrete case study and visualization method, we demonstrate how \alg offers an intuitive interpretation.



\section{Methodology}
\label{sec:method}
We first describe the structure of sequential EHR data and our notation, then follow with a general framework for predictive analysis in healthcare using EHR, followed by details of the \alg method.

\textbf{EHR Structure and our Notation.} The EHR data of each patient can be represented as a time-labeled sequence of multivariate observations. Assuming we use $r$ different variables, the $n$-th patient of $N$ total patients can be represented by a sequence of $T^{(n)}$ tuples $(t_i^{(n)}, \mathbf{x}_i^{(n)}) \in \mathbb{R}\times\mathbb{R}^{r} , i=1, \ldots, T^{(n)}$. The timestamps $t_i^{(n)}$ denotes the time of the $i$-th visit of the $n$-th patient and $T^{(n)}$ the number of visits of the $n$-th patient. To minimize clutter, we describe the algorithms for a single patient and have dropped the superscript $(n)$ whenever it is unambiguous. The goal of predictive modeling is to predict the label at each time step $\mathbf{y}_i \in \{0,1\}^{s}$ or at the end of the sequence $\mathbf{y} \in \{0,1\}^{s}$. The number of labels $s$ can be more than one.

For example, in encounter sequence modeling (ESM) \cite{choi2015doctor}, each visit (\textit{e.g.} encounter) of a patient's visit sequence is represented by a set of varying number of medical codes $\{c_1, c_2, \ldots, c_n\}$. $c_j$ is the $j$-th code from the vocabulary $\mathcal{C}$. Therefore, in ESM, the number of variables $r=|\mathcal{C}|$ and input $\xb_i \in \{0,1\}^{|\mathcal{C}|}$ is a binary vector where the value one in the $j$-th coordinate indicates that $c_j$ was documented in $i$-th visit. Given a sequence of visits $\xb_1, \ldots, \xb_T$, the goal of ESM is, for each time step $i$, to predict the codes occurring at the next visit $\xb_2, \ldots, \xb_{T+1}$, with the number of labels $s=|\mathcal{C}|$.

In case of learning to diagnose (L2D) \cite{lipton2015learning}, the input vector $\xb_i$ consists of continuous clinical measures. If there are $r$ different measurements, then $\xb_i \in \mathbb{R}^{r}$. The goal of L2D is, given an input sequence $\xb_1, \ldots, \xb_T$, to predict the occurrence of a specific disease ($s=1$) or multiple diseases ($s > 1$). 
Without loss of generality, we will describe the algorithm for ESM, as L2D can be seen as a special case of ESM where we make a single prediction at the end of the visit sequence.

In the rest of this section, we will use the abstract symbol $\mathrm{RNN}$ to denote any recurrent neural network variants that can cope with the vanishing gradient problem \cite{bengio1994learning}, such as LSTM \cite{hochreiter1997long}, GRU \cite{cho2014learning}, and IRNN \cite{le2015simple}, with any depth (number of hidden layers). 

\subsection{Preliminaries on Neural Attention Models}
\label{sec:attention}
Attention based neural network models are being successfully applied to image processing \cite{ba2014multiple,mnih2014recurrent,gregor2015draw,xu2015show}, natural language processing \cite{bahdanau2014neural,hermann2015teaching,rush2015neural} and speech recognition \cite{chorowski2015attention}. The utility of the attention mechanism can be seen in the language translation task \cite{bahdanau2014neural} where it is inefficient to represent an entire sentence with one fixed-size vector because neural translation machines finds it difficult to translate the given sentence represented by a single vector. 

Intuitively, the attention mechanism for language translation works as follows: given a sentence of length $S$ in the original language, we generate $\mathbf{h}_1, \ldots, \mathbf{h}_{S}$, to represent the words in the sentence. To find the $j$-th word in the target language, we generate attentions $\alpha_i^j$ for $i=1, \ldots, S$ for each word in the original sentence. Then, we compute the context vector $\mathbf{c}_j = \sum_{i}\alpha_i^j\mathbf{h}_i$ and use it to predict the $j$-th word in the target language.  In general, the attention mechanism allows the model to focus on a specific word (or words) in the given sentence when generating each word in the target language. 

We rely on a conceptually similar temporal attention mechanism to generate interpretable prediction models using EHR data. Our model framework is motivated by and mimics how doctors attend to a patient’s needs and explore the patient record, where there is a focus on specific clinical information (e.g., key risk factors) working from the present to the past.

 
\subsection{Reverse Time Attention Model \alg}
\label{ssec:model}
\begin{figure}[t]
\centering
\includegraphics[scale=0.25]{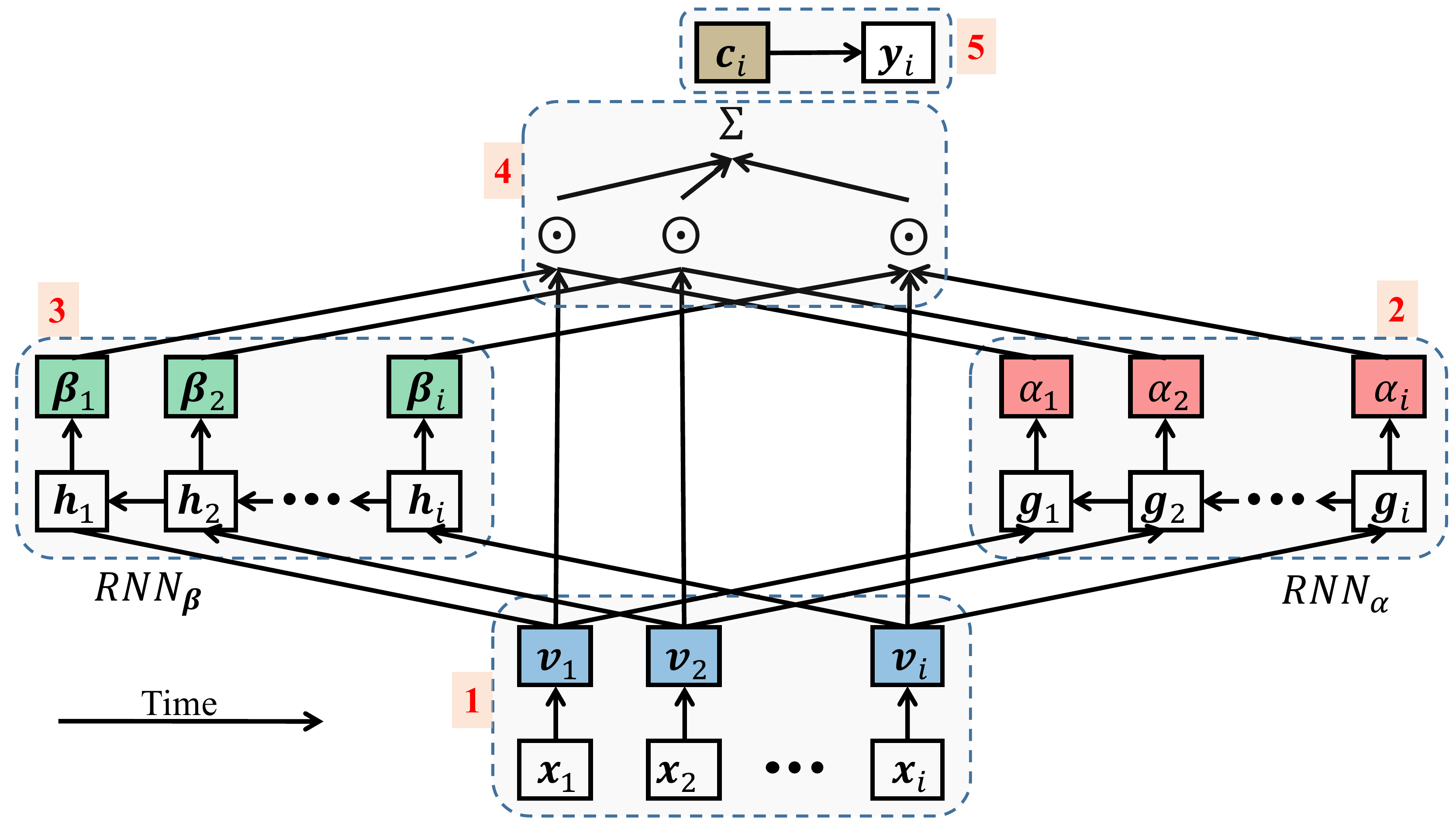}
\caption{Unfolded view of \alg's architecture: Given input sequence $\xb_1, \ldots, \xb_i$, we predict the label $\yb_i$. 
\textbf{Step} $\bm{1}$: Embedding, 
\textbf{Step} $\bm{2}$: generating $\alpha$ values using $\mathrm{RNN}_{\alpha}$, 
\textbf{Step} $\bm{3}$: generating $\bm{\beta}$ values using $\mathrm{RNN}_{\bm{\beta}}$, 
\textbf{Step} $\bm{4}$: Generating the context vector using attention and representation vectors, and 
\textbf{Step} $\bm{5}$: Making prediction.
Note that in Steps 2 and 3 we use RNN in the reversed time.}
\label{fig:architecture}
\end{figure}

Figure \ref{fig:architecture} shows the high-level overview of our model, where a central feature is to delegate a considerable portion of the prediction responsibility to the process for generating attention weights. This is intended to address, in part, the difficulty with interpreting RNNs where the recurrent weights feed past information to the hidden layer. Therefore, to consider both the visit-level and the variable-level (individual coordinates of $\xb_i$) influence, we use a linear embedding of the input vector $\xb_i$.  That is, we define
\begin{equation}
\vb_i = \Wb_{emb} \xb_i,  \tag{Step 1}
\end{equation}
\noindent where $\vb_i \in \mathbb{R}^{m}$ denotes the embedding of the input vector $\xb_i \in \mathbb{R}^{r}$, $m$ the size of the embedding dimension, $\Wb_{emb} \in \mathbb{R}^{m \times r}$ the embedding matrix to learn. We can alternatively use more sophisticated yet interpretable representations such as those derived from multilayer perceptron (MLP)~\cite{erhan2009visualizing,le2013building}. MLP has been used for representation learning in EHR data \cite{choi2016multi}.

We use two sets of weights, one for the visit-level attention and the other for variable-level attention, respectively. The scalars $\alpha_1, \ldots, \alpha_i$ are the visit-level attention weights that govern the influence of each visit embedding $\vb_1, \ldots, \vb_i$. The vectors $\bm{\beta}_1, \ldots, \bm{\beta}_i$ are the variable-level attention weights that focus on each coordinate of the visit embedding $v_{1,1}, v_{1,2}, \ldots, v_{1,m}, \ldots, v_{i,1}, v_{i,2}, \ldots, v_{i,m}$. 

We use two RNNs, $\mathrm{RNN}_{\alpha}$ and $\mathrm{RNN}_{\bm{\beta}}$, to separately generate $\alpha$'s and $\bm{\beta}$'s as follows,
\begin{align}
\gb_i, \gb_{i-1}, \ldots, \gb_1 &= \mathrm{RNN}_{\alpha}(\vb_i, \vb_{i-1}, \ldots, \vb_1), \nonumber \\
e_j &= \wb_{\alpha}^{\top} \gb_j +  b_{\alpha}, \quad \text{for} \quad j = 1, \ldots, i \label{eq:alpha} \nonumber \\
\alpha_1, \alpha_2, \ldots, \alpha_i &= \soft(e_1, e_2, \ldots, e_i) \tag{Step 2} \nonumber\\
\hb_i, \hb_{i-1}, \ldots, \hb_1 &= \mathrm{RNN}_{\bm{\beta}}(\vb_i, \vb_{i-1}, \ldots, \vb_1) \nonumber \\
\bm{\beta}_j &= \tanh\big(\Wb_{\bm{\beta}}\hb_j + \bb_{\bm{\beta}}\big)  \quad \text{for}\quad j = 1, \ldots, i, \tag{Step 3}\label{eq:beta}
\end{align}
where $\gb_i \in \mathbb{R}^{p}$ is the hidden layer of $\mathrm{RNN}_{\alpha}$ at time step $i$, $\hb_i \in \mathbb{R}^{q}$ the hidden layer of $\mathrm{RNN}_{\bm{\beta}}$ at time step $i$ and $\wb_{\alpha} \in \mathbb{R}^{p}, b_{\alpha} \in \mathbb{R}, \Wb_{\bm{\beta}} \in \mathbb{R}^{m \times q}$ and $\bb_{\bm{\beta}} \in \mathbb{R}^{m}$ are the parameters to learn. The hyperparameters $p$ and $q$ determine the hidden layer size of $\mathrm{RNN}_{\alpha}$ and $\mathrm{RNN}_{\bm{\beta}}$, respectively. 
Note that for prediction at each timestamp, we generate a new set of attention vectors $\alpha$ and $\bm{\beta}$. For simplicity of notation, we do not include the index for predicting at different time steps. 
In \ref{eq:alpha}, we can use Sparsemax \cite{martins2016softmax} instead of Softmax for sparser attention weights.

As noted, \alg generates the attention vectors by running the RNNs backward in time;  i.e., $\mathrm{RNN}_{\alpha}$ and $\mathrm{RNN}_{\bm{\beta}}$ both take the visit embeddings in a reverse order $\vb_i, \vb_{i-1}, \ldots, \vb_1$. Running the RNN in reversed time order also offers computational advantages since the reverse time order allows us to generate $e$'s and $\bm{\beta}$'s that dynamically change their values when making predictions at different time steps $i=1, 2, \ldots, T$.
This ensures that the attention vectors are modified at each time step, increasing the computational stability of the attention generation process.\footnote{For example, feeding visit embeddings in the original order to $\mathrm{RNN}_{\alpha}$ and $\mathrm{RNN}_{\bm{\beta}}$ will generate the same $e_1$ and $\bm{\beta}_1$ for every time step $i=1, 2, \ldots, T$. Moreover, in many cases, a patient's recent visit records deserve more attention than the old records. Then we need to have $e_{j+1}>e_{j}$ which makes the process computationally unstable for long sequences.} 

Using the generated attentions, we obtain the context vector $\cbb_i$ for a patient up to the $i$-th visit as follows,
\begin{equation}
\cbb_i = \sum_{j=1}^{i} \alpha_j \bm{\beta}_j \odot \vb_j, \tag{Step 4} \label{eq:context}
\end{equation}
where $\odot$ denotes element-wise multiplication. We use the context vector $\cbb_i \in \mathbb{R}^{m}$ to predict the true label $\yb_i \in \{0,1\}^{s}$ as follows,
\begin{equation}
\widehat{\yb}_{i} = \mathrm{Softmax}(\Wb\cbb_i + \bb), \label{eq:softmax} \tag{Step 5}
\end{equation}
where $\Wb \in \mathbb{R}^{s \times m}$ and $\bb \in \mathbb{R}^{s}$ are parameters to learn. We use the cross-entropy to calculate the classification loss as follows, 
\begin{equation} 
\mathcal{L}(\xb_1, \ldots, \xb_T) = -\frac{1}{N}\sum_{n=1}^{N} \frac{1}{T^{(n)}} \sum_{i=1}^{T^{(n)}} \Big( \yb_{i}^{\top} \log(\widehat{\yb}_{i}) + (\mathbf{1} - \yb_{i})^{\top}  \log(\mathbf{1} - \widehat{\yb}_{i})  \Big) \label{eq:cross_entropy}
\end{equation}
where we sum the cross entropy errors from all dimensions of $\widehat{\yb}_{i}$. In case of real-valued output $\yb_i \in \mathbb{R}^{s}$, we can change the cross-entropy in Eq.~\eqref{eq:cross_entropy} to, for example, mean squared error.

Overall, our attention mechanism can be viewed as the inverted architecture of the standard attention mechanism for NLP \cite{bahdanau2014neural} where the words are encoded by RNN and the attention weights are generated by MLP. In contrast, our method uses MLP to embed the visit information to preserve interpretability and uses RNN to generate two sets of attention weights, recovering the sequential information as well as mimicking the behavior of physicians. 
Note that we did not use the timestamp of each visit in our formulation. Using timestamps, however, provides a small improvement in the prediction performance. We propose a method to use timestamps in	 Appendix \ref{sec:timestamp}.

\section{Interpreting \alg}
\label{sec:interpretation}
Finding the visits that contribute to prediction are derived using the largest $\alpha_i$, which is straightforward. 
However, finding influential variables is slightly more involved as a visit is represented by an ensemble of medical variables, each of which can vary in its predictive contribution.
The contribution of each variable is determined by $\vb$, $\bm{\beta}$ and $\alpha$, and interpretation of $\alpha$ alone informs which visit is influential in prediction but not why. 

We propose a method to interpret the end-to-end behavior of \alg. By keeping $\alpha$ and $\bm{\beta}$ values fixed as the attention of doctors, we analyze changes in the probability of each label $y_{i,1}, \ldots, y_{i,s}$ in relation to changes in the original input  $x_{1,1}, \ldots, x_{1,r}, \ldots, x_{i,1}, \ldots, x_{i,r}$. The $x_{j,k}$ that yields the largest change in $y_{i,d}$ will be the input variable with highest contribution.
More formally, given the sequence $\xb_1, \ldots, \xb_i$, we are trying to predict the probability of the output vector $\yb_i \in \{0,1\}^{s}$, which can be expressed as follows 
\begin{equation}
p(\yb_{i} | \xb_1, \ldots, \xb_i) = p(\yb_{i} | \cbb_i) = \soft \left(\Wb\cbb_i + \bb\right) \label{eq:di_prob}
\end{equation}  
where $\cbb_i \in \mathbb{R}^{m}$ denotes the context vector. According to \ref{eq:context}, $\cbb_i$ is the sum of the visit embeddings $\vb_1, \ldots, \vb_i$ weighted by the attentions $\alpha$'s and $\bm{\beta}$'s. Therefore Eq~\eqref{eq:di_prob} can be rewritten as follows, 
\begin{equation}
p(\yb_{i} | \xb_1, \ldots, \xb_i) = p(\yb_{i} | \cbb_i) = \soft\bigg( \Wb \Big( \sum_{j=1}^{i} \alpha_j \bm{\beta}_j \odot \vb_j \Big) + \bb \bigg) \label{eq:di_prob_decom1}
\end{equation}
Using the fact that the visit embedding $\vb_i$ is the sum of the columns of $\Wb_{emb}$ weighted by each element of $\xb_i$, Eq~\eqref{eq:di_prob_decom1} can be rewritten as follows, 
\begin{align}
p(\yb_{i} | \xb_1, \ldots, \xb_i) & = \soft \bigg( \Wb \Big( \sum_{j=1}^{i} \alpha_j \bm{\beta}_j \odot \sum_{k=1}^{r} x_{j,k} \Wb_{emb}[:,k] \Big) + \bb \bigg) \nonumber \\
& = \soft \bigg(\sum_{j=1}^{i} \sum_{k=1}^{r} x_{j,k} \, \alpha_j  \Wb \Big( \bm{\beta}_j \odot \Wb_{emb}[:,k] \Big) + \bb \bigg) \label{eq:di_prob_decom2}
\end{align}
where $x_{j,k}$ is the $k$-th element of the input vector $\xb_j$. Eq~\eqref{eq:di_prob_decom2} can be completely deconstructed to the variables at each input $\xb_1, \ldots, \xb_i$, which allows for calculating the contribution $\omega$ of the $k$-th variable of the input $\xb_j$ at time step $j \le i$, for predicting $\yb_i$ as follows, 
\begin{equation}
\omega(\yb_{i}, x_{j,k}) = \underbrace{\alpha_j \Wb ( \bm{\beta}_j \odot \Wb_{emb}[:,k] )}_{\text{Contribution coefficient}} \underbrace{x_{j,k}}_{\text{Input value}},  \label{eq:contribution}
\end{equation}
where the index $i$ of $\yb_i$ is omitted in the $\alpha_j$ and $\bm{\beta}_j$. As we have described in Section \ref{ssec:model}, we are generating $\alpha$'s and $\bm{\beta}$'s at time step $i$ in the visit sequence $\xb_1, \ldots, \xb_T$. Therefore the index $i$ is always assumed for $\alpha$'s and $\bm{\beta}$'s. Additionally, Eq~\eqref{eq:contribution} shows that when we are using a binary input value, the coefficient itself is the contribution. However, when we are using a non-binary input value, we need to multiply the coefficient and the input value $x_{j,k}$ to correctly calculate the contribution. 


\section{Experiments}
\label{sec:experiments}
We compared performance of \alg to RNNs and traditional machine learning methods. Given space constraints, we only report the results on the learning to diagnose (L2D) task and summarize the encounter sequence modeling (ESM) in Appendix \ref{sec:dpm}. The \alg source code is publicly available at \url{https://github.com/mp2893/retain}.

\vspace{-0.1in}
\subsection{Experimental setting}
\vspace{-0.1in}
\label{sec:settings}
\textbf{Source of data:} The dataset consists of electronic health records from Sutter Health. The patients are 50 to 80 years old adults chosen for a heart failure prediction model study. From the encounter records, medication orders, procedure orders and problem lists, we extracted visit records consisting of diagnosis, medication and procedure codes. To reduce the dimensionality while preserving the clinical information, we used existing medical groupers to aggregate the codes into input variables. The details of the medical groupers are given in the Appendix \ref{sec:supp}. A profile of the dataset is summarized in Table \ref{table:data}.

\begin{table}[t]
\caption{Statistics of EHR dataset. (D:Diagnosis, R:Medication, P:Procedure)}
\label{table:data}
\centering
\begin{tabular}{l|c||l|c}
\hline
\# of patients & 263,683 & Avg. \# of codes in a visit & 3.03 \\
\# of visits & 14,366,030 & Max \# of codes in a visit & 62 \\
Avg. \# of visits per patient & 54.48 & Avg. \# of Dx codes in a visit & 1.83\\
\# of medical code groups  & 615 (D:283, R:94, P:238) & Max \# of Dx  in a visit & 42 \\
\hline
\end{tabular}
\end{table}

\textbf{Implementation details:} We implemented \alg with Theano 0.8 \cite{bergstra2010theano}. For training the model, we used Adadelta \cite{zeiler2012adadelta} with the mini-batch of 100 patients. The training was done in a machine equipped with Intel Xeon E5-2630, 256GB RAM, two Nvidia Tesla K80's and CUDA 7.5. 

\textbf{Baselines:} For comparison, we completed the following models. 

\vspace{-0.05in}
\begin{itemize}[leftmargin=5.5mm]
\item
\textbf{Logistic regression (LR)}: We compute the counts of  medical codes for each patient based on all her visits as input variables and normalize the vector to zero mean and unit variance. We use the resulting vector to train the logistic regression. 
\vspace{-0.05in}
\item
\textbf{MLP}: We use the same feature construction as \textbf{LR}, but put a hidden layer of size 256 between the input and output.
\vspace{-0.05in}
\item
\textbf{RNN}: RNN with two hidden layers of size 256 implemented by the GRU. Input sequences $\xb_1, \ldots, \xb_i$ are used. Logistic regression is applied to the top hidden layer. We use two layers of RNN of to match the model complexity of \alg.
\vspace{-0.05in}
\item
\textbf{RNN+$\bm{\alpha}_{M}$}: One layer single directional RNN (hidden layer size 256) along time to generate the input embeddings $\vb_1, \ldots, \vb_i$. We use the MLP with a single hidden layer of size 256 to generate the visit-level attentions $\alpha_1, \ldots, \alpha_i$. We use the input embeddings $\vb_1, \ldots, \vb_i$ as the input to the MLP. This baseline corresponds to Figure \ref{fig:common}.
\vspace{-0.05in}
\item
\textbf{RNN+$\bm{\alpha}_{R}$}: This is similar to \textbf{RNN+$\bm{\alpha}_{M}$} but uses the reverse-order RNN (hidden layer size 256) to generate the visit-level attentions $\alpha_1, \ldots, \alpha_i$. We use this baseline to confirm the effectiveness of generating the attentions using reverse time order.
\end{itemize}
\vspace{-0.05in}
The comparative visualization of the baselines are provided in Appendix \ref{sec:baseFigs}. We use the same implementation and training method for the baselines as described above. The details on the hyper-parameters, regularization and drop-out strategies for the baselines are described in Appendix \ref{sec:supp}.

\textbf{Evaluation measures:} Model accuracy was measured by:
\begin{itemize}[leftmargin=5.5mm]
\item
\textbf{Negative log-likelihood} that measures the model loss on the test set. The loss can be calculated by Eq~\eqref{eq:cross_entropy}.
\vspace{-0.5mm}
\item 
\textbf{Area Under the ROC Curve (AUC)} of comparing $\widehat{y}_i$ with the true label $y_i$. 
AUC is more robust to imbalanced positive/negative prediction labels, making it appropriate for evaluation of classification accuracy in the heart failure prediction task.
\end{itemize}
We also report the bootstrap (10,000 runs) estimate of the standard deviation of the evaluation measures.



\subsection{Heart Failure Prediction}
\textbf{Objective:} Given a visit sequence $\xb_1, \ldots, \xb_T$, we predicted if a primary care patient will be diagnosed with heart failure (HF). This is a special case of ESM with a single disease outcome at the end of the sequence. Since this is a binary prediction task, we use the logistic sigmoid function instead of the Softmax in \ref{eq:softmax}.

\textbf{Cohort construction:} From the source dataset, 3,884 cases are selected and approximately 10 controls are selected for each case (28,903 controls). The case/control selection criteria are fully described in the supplementary section. Cases have index dates to denote the date they are diagnosed with HF. Controls have the same index dates as their corresponding cases. We extract diagnosis codes, medication codes and procedure codes in the 18-months window before the index date. 

\textbf{Training details:} The patient cohort was divided into the training, validation and test sets in a 0.75:0.1:0.15 ratio. The validation set was used to determine the values of the hyper-parameters. See Appendix \ref{sec:supp} for details of hyper-parameter tuning.



\begin{table}[t]
\centering
    \caption{Heart failure prediction performance of \alg and the baselines}
    \begin{tabular}{l|c|c||c|c}
      {Model} & {Test Neg Log Likelihood} & { AUC} & Train Time / epoch & Test Time\\
      \hline \hline
      LR & $0.3269\pm0.0105$ & $0.7900\pm0.0111$ & 0.15s & 0.11s \\

      MLP & $0.2959\pm0.0083$ & $0.8256\pm0.0096$ & 0.25s & 0.11s \\
      \hline
      RNN & $0.2577\pm0.0082$ & $0.8706\pm0.0080$ & 10.3s & 0.57s \\
      \hline
      RNN+$\alpha_M$ & $0.2691\pm0.0082$ & $0.8624\pm0.0079$ & 6.7s & 0.48s\\
      RNN+$\alpha_R$ & $0.2605\pm0.0088$ & $\mathbf{0.8717}\pm0.0080$ & 10.4s & 0.62s \\
      \hline
      \alg & $\mathbf{0.2562}\pm0.0083$ & $0.8705\pm0.0081$ & 10.8s & 0.63s \\
            \hline
    \end{tabular}
\label{tab:hfResult}
\vspace{-0.1in}
\end{table} 
\textbf{Results:} Logistic regression and MLP underperformed compared to the four temporal learning algorithms (Table \ref{tab:hfResult}). \alg is comparable to the other RNN variants in terms of prediction performance while offering the interpretation benefit. 

Note that RNN+$\alpha_R$ model  are a degenerated version of \alg with only scalar attention, which is still a competitive model as shown in table \ref{tab:hfResult}. This confirms the efficiency of generating attention weights using the RNN. 
However, RNN+$\alpha_R$ model only provides scalar  visit-level attention, which is not sufficient for healthcare applications. 
Patients often receives several medical codes at a single visit, and it will be important to distinguish their relative importance to the target. We show such a case study in section \ref{ssec:analysis}.

Table \ref{tab:hfResult} also shows the scalability of \alg, as its training time (the number of seconds to train the model over the entire training set once) is comparable to RNN. The test time is the number of seconds to generate the prediction output for the entire test set. We use the mini-batch of 100 patients when assessing both training and test times. RNN takes longer than RNN+$\alpha_M$ because of its two-layer structure, whereas RNN+$\alpha_M$ uses a single layer RNN.
The models that use two RNNs (RNN, RNN+$\alpha_R$, \alg)\footnote{The RNN baseline uses two layers of $\mathrm{RNN}$, RNN+$\alpha_R$ uses one for visit embedding and one for generating $\alpha$, \alg uses each for generating $\alpha$ and $\bm{\beta}$} take similar time to train for one epoch. However, each model required a different number of epochs to converge. RNN typically takes approximately 10 epochs, RNN+$\alpha_M$ and RNN+$\alpha_R$ 15 epochs and \alg 30 epochs. 
Lastly, training the attention models (RNN+$\alpha_M$, RNN+$\alpha_R$ and \alg) for ESM would take considerably longer than L2D, because ESM modeling generates context vectors at each time step. RNN, on the other hand, does not require additional computation other than embedding the visit to its hidden layer to predict target labels at each time step. Therefore, in ESM, the training time of the attention models will increase linearly in relation to the length of the input sequence. 

\vspace{-0.1in}
\subsection{Model Interpretation for Heart Failure Prediction}
\begin{figure}[t]
\includegraphics[width=\textwidth]{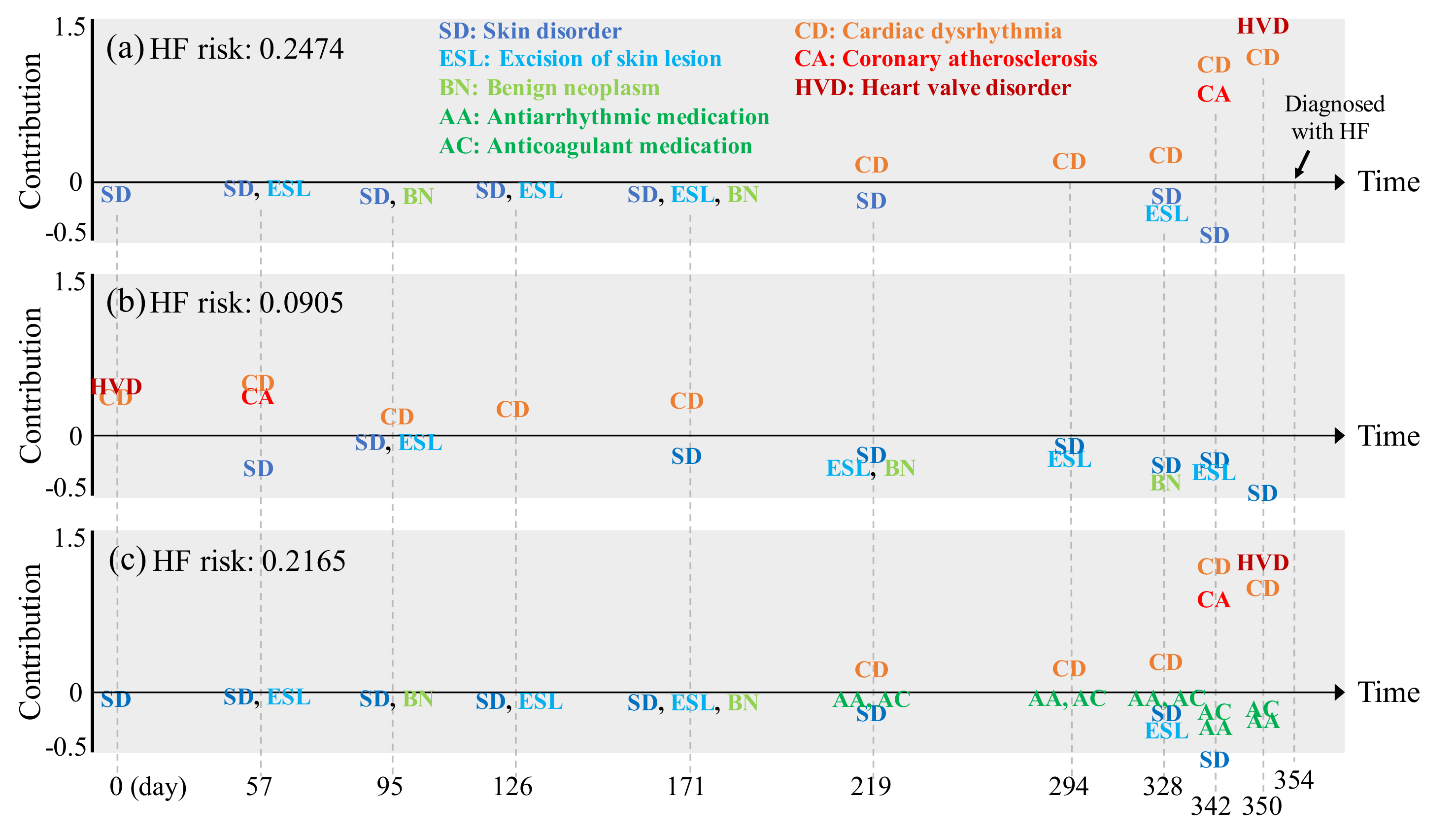}
\caption{(a) Temporal visualization of a patient's visit records where the contribution of variables for diagnosis of heart failure (HF) is summarized along the $x$-axis (\textit{i.e.} time) with the $y$-axis indicating the magnitude of visit and code specific contributions to HF diagnosis. (b) We reverse the order of the visit sequence to see if \alg can properly take into account the modified sequence information. (c) Medication codes are added to the visit record to see how it changes the behavior of \alg.}
\label{fig:visualization}
\vspace{-0.2in}
\end{figure}

\vspace{-0.1in}
\label{ssec:analysis}
We evaluated the interpretability of \alg in the HF prediction task by choosing a HF patient from the test set and calculating the contribution of the variables (medical codes in this case) to diagnostic prediction. The patient suffered from skin problems, \textit{skin disorder} (SD), \textit{benign neoplasm} (BN), \textit{excision of skin lesion} (ESL), for some time before showing symptoms of HF, \textit{cardiac dysrhythmia} (CD), \textit{heart valve disease} (HVD) and \textit{coronary atherosclerosis} (CA), and then a diagnosis of HF (Figure \ref{fig:visualization}). We can see that skin-related codes from the earlier visits made little contribution to HF prediction as expected. 
\alg properly puts much attention to the HF-related codes that occurred in recent visits. 

To confirm \alg's ability to exploit the sequence information of the EHR data, we reverse the visit sequence of Figure \ref{fig:visualization}a and feed it to \alg. 
Figure \ref{fig:visualization}b shows the contribution of the medical codes of the reversed visit record. HF-related codes in the past are still making positive contributions, but not as much as they did in Figure \ref{fig:visualization}a. 
Figure \ref{fig:visualization}b also emphasizes \alg's superiority to interpretable, but stationary models such as logistic regression. Stationary models often aggregate past information and remove the temporality from the input data, which can mistakenly lead to the same risk prediction for Figure \ref{fig:visualization}a and \ref{fig:visualization}b. \alg, however, can correctly digest the sequence information and calculates the HF risk score of 9.0\%, which is significantly lower than that of Figure \ref{fig:visualization}a.

Figure \ref{fig:visualization}c shows how the contributions of codes change when selected medication data are used in the model. We added two medications from day 219: \textit{antiarrhythmics} (AA) and \textit{anticoagulants} (AC), both of which are used to treat \textit{cardiac dysrhythmia} (CD). The two medications make a negative contributions, especially towards the end of the record. The medications decreased the positive contributions of \textit{heart valve disease} and \textit{cardiac dysrhythmia} in the last visit. Indeed, the HF risk prediction (0.2165) of Figure \ref{fig:visualization}c is lower than that of Figure \ref{fig:visualization}a (0.2474). This suggests that taking proper medications can help the patient in reducing their HF risk.
\vspace{-0.1in}

\section{Conclusion}
\label{sec:conclusion}

Our approach to modeling event sequences as predictors of HF diagnosis suggest that complex models can offer both superior predictive accuracy and more precise interpretability.
Given the power of RNNs for analyzing sequential data, we proposed \alg, which preserves RNN's predictive power while allowing a higher degree of interpretation.
The key idea of \alg is to improve the prediction accuracy through a sophisticated attention generation process, while keeping the representation learning part simple for interpretation, making the entire algorithm accurate and interpretable. 
\alg trains two RNN in a reverse time order to efficiently generate the appropriate attention variables. 
For future work, we plan to develop an interactive visualization system for \alg and evaluating \alg in other healthcare applications.

\footnotesize
\bibliography{nips2016}
\bibliographystyle{abbrv}
\normalsize

\newpage
\clearpage
\appendix
\section{A method to use the timestamps}
\label{sec:timestamp}
As before, we use $t_{i}^{(n)}$ to represent the timestamp of the $i$-th visit of the $n$-th patient. In the following, we suppress the superscript $(n)$ to avoid cluttered notation. Note that the timestamp $t_i$ can be anything that provides the temporal information of the $i$-th visit: the number of days from the first visit, the number of days between two consecutive visits, or the number of days until the index date of some event such as heart failure diagnosis.

In order to use the timestamps, we modify Step 2 and Step 3 in Section \ref{ssec:model} as follows:
\begin{align}
\gb_i, \gb_{i-1}, \ldots, \gb_1 &= \mathrm{RNN}_{\alpha}(\vb_{i}^{'}, \vb_{i-1}^{'}, \ldots, \vb_{1}^{'}), \nonumber \\
e_j &= \wb_{\alpha}^{\top} \gb_j +  b_{\alpha}, \quad \text{for} \quad j = 1, \ldots, i \nonumber \\
\alpha_1, \alpha_2, \ldots, \alpha_i &= \soft(e_1, e_2, \ldots, e_i) \nonumber\\
\hb_i, \hb_{i-1}, \ldots, \hb_1 &= \mathrm{RNN}_{\bm{\beta}}(\vb_{i}^{'}, \vb_{i-1}^{'}, \ldots, \vb_{1}^{'}) \nonumber \\
\bm{\beta}_j &= \tanh\big(\Wb_{\bm{\beta}}\hb_j + \bb_{\bm{\beta}}\big)  \quad \text{for}\quad j = 1, \ldots, i, \nonumber \\
\mbox{where } \vb_{i}^{'} &= [\vb_i, t_i]\nonumber
\end{align}
where we use $\vb_{i}^{'}$, the concatenation of the visit embedding $\vb_i$ and the timestamp $t_i$, to generate the attentions $\alpha$ and $\bm{\beta}$. However, when obtaining the context vector $\cbb_i$ as per Step 4, we use $\vb_i$, not $\vb_{i}^{'}$ to match the dimensionality. The entire process could be understood such that we use the temporal information not to embed each visit, but to calculate the attentions for the entire visit sequence. This is consistent with our modeling approach where we lose the sequential information in embedding the visit with MLP, then recover the sequential information by generating the attentions using the RNN. By using the temporal information, specifically the log of the number of days from the first visit, we were able to improve the heart failure prediction AUC by 0.003 without any hyper-parameter tuning.

\section{Details of the experiment settings}
\label{sec:supp}

\subsection{Hyper-parameter Tuning}
We used the validation set to tune the hyper-parameters: visit embedding size $m$, $\mathrm{RNN}_{\alpha}$'s hidden layer size $p$, $\mathrm{RNN}_{\bm{\beta}}$'s hidden layer size $q$, $L_2$ regularization coefficient, and drop-out rates. 

$L_2$ regularization was applied to all weights except the ones in $\mathrm{RNN}_{\alpha}$ and $\mathrm{RNN}_{\bm{\beta}}$. Two separate drop-outs were used on the visit embedding $\vb_i$ and the context vector $\cbb_i$. We performed the random search with predefined ranges $m,p,q \in \{32,64,128,200,256\}$, $L_2 \in \{0.1,0.01,0.001,0.0001\}$, $dropout_{\vb_i}, dropout_{\cbb_i} \in \{0.0,0.2,0.4,0.6,0.8\}$. We also performed the random search with $m$, $p$ and $q$ fixed to 256.

The final value we used to train \alg for heart failure prediction is $m,p,q=128$, $dropout_{\vb_i}=0.6$, $dropout_{\cbb_i}=0.6$ and 0.0001 for the $L_2$ regularization coefficient. 

\subsection{Code Grouper}
Diagnosis codes, medication codes and procedure codes in the dataset are respectively using International Classification of Diseases (ICD-9), Generic Product Identifier (GPI) and Current Procedural Terminology (CPT). 

Diagnosis codes are grouped by Clinical Classifications Software for ICD-9\footnote{https://www.hcup-us.ahrq.gov/toolssoftware/ccs/ccs.jsp} which reduces the number of diagnosis code from approximately 14,000 to 283. Medication codes are grouped by Generic Product Identifier Drug Group\footnote{http://www.wolterskluwercdi.com/drug-data/medi-span-electronic-drug-file/} which reduces the dimension to from approximately 151,000 to 96. Procedure codes are grouped by Clinical Classifications Software for CPT\footnote{https://www.hcup-us.ahrq.gov/toolssoftware/ccs\_svcsproc/ccssvcproc.jsp}, which reduces the number of CPT codes from approximately 9,000 to 238. 

\subsection{Training Specifics of the Basline Models}
\begin{itemize}[leftmargin=5.5mm]
\item
\textbf{LR}: We use 0.01 $L_2$ regularization coefficient for the logistic regression weight.
\item
\textbf{MLP}: We use drop-out rate 0.6 on the output of the hidden layer. We use 0.0001 $L_2$ regularization coefficient for the hidden layer weight and the logistic regression weight.
\item
\textbf{RNN}: We use drop-out rate 0.6 on the outputs of both hidden layers. We use 0.0001 $L_2$ regularization coefficient for the logistic regression weight. The dimension size of both hidden layers is 256.
\item
\textbf{RNN+$\bm{\alpha}_{M}$}: We use drop-out rate 0.4 on the output of the hidden layer and 0.6 on the output of the context vector $\sum_i \alpha_i \vb_i$. We use 0.0001 $L_2$ regularization coefficient for the hidden layer weight of the MLP that generates $\alpha$'s and the logistic regression weight. The dimension size of the hidden layers in both RNN and MLP is 256.
\item
\textbf{RNN+$\bm{\alpha}_{R}$}: We use drop-out rate 0.4 on the output of the hidden layer and 0.6 on the output of the context vector $\sum_i \alpha_i \vb_i$. We use 0.0001 $L_2$ regularization coefficient for the hidden layer weight of the RNN that generates $\alpha$'s and the logistic regression weight. The dimension size of the hidden layers in both RNNs is 256.
\end{itemize}

\subsection{Heart Failure Case/Control Selection Criteria}
\begin{figure}
\centering
\captionof{table}{Qualifying ICD-9 codes for heart failure}
\includegraphics[scale=0.6]{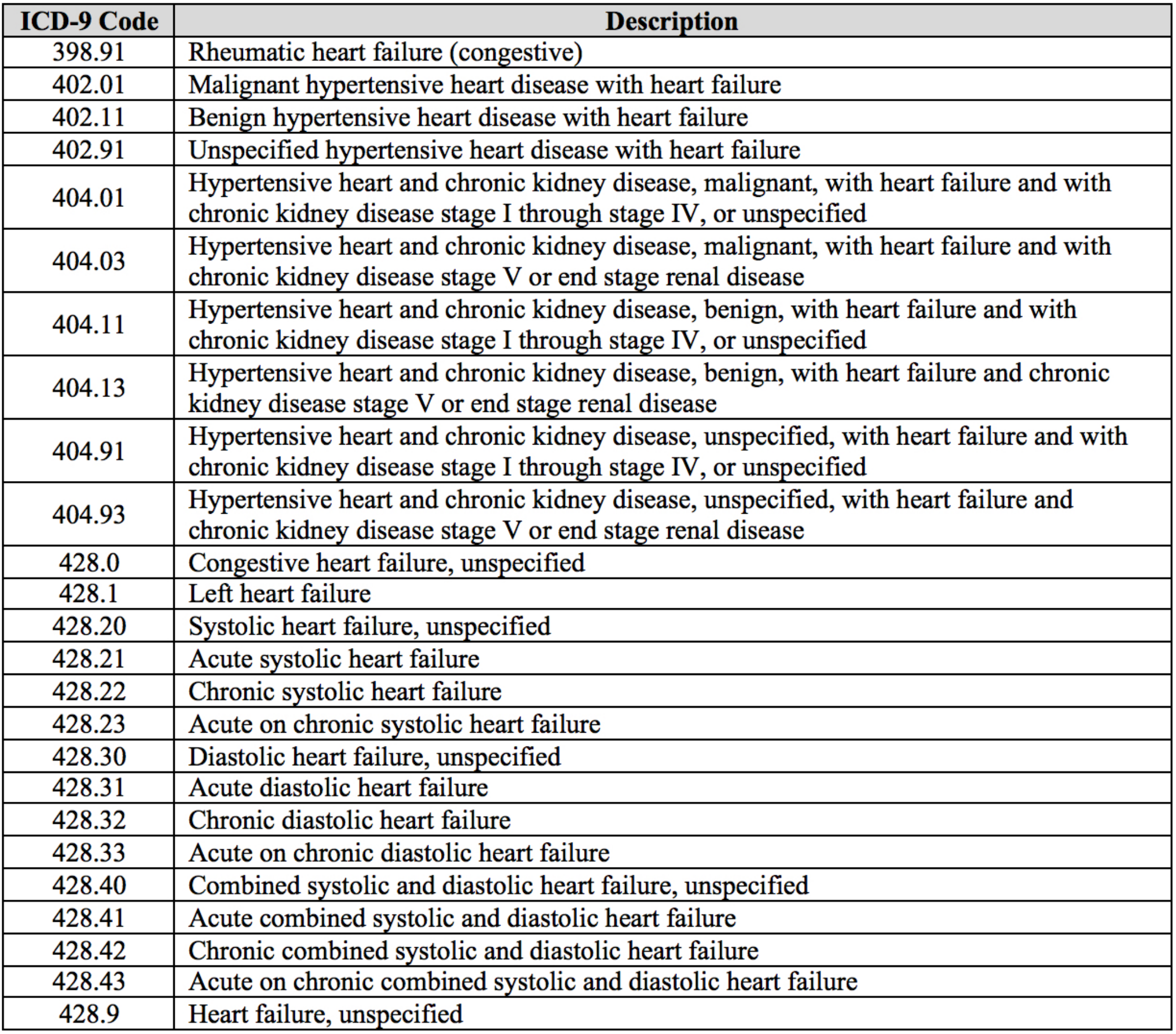}
\label{table:hf_codes}
\end{figure}
Case patients were 40 to 85 years of age at the time of HF diagnosis. HF diagnosis (HFDx) is defined as: 1) Qualifying ICD-9 codes for HF appeared in the encounter records or medication orders. Qualifying ICD-9 codes are displayed in Table \ref{table:hf_codes}. 2) a minimum of three clinical encounters with qualifying ICD-9 codes had to occur within 12 months of each other, where the date of diagnosis was assigned to the earliest of the three dates.  If the time span between the first and second appearances of the HF diagnostic code was greater than 12 months, the date of the second encounter was used as the first qualifying encounter.  The date at which HF diagnosis was given to the case is denoted as HFDx.
Up to ten eligible controls (in terms of sex, age, location) were selected for each case, yielding an overall ratio of 9 controls per case. Each control was also assigned an index date, which is the HFDx of the matched case. Controls are selected such that they did not meet the operational criteria for HF diagnosis prior to the HFDx plus 182 days of their corresponding case. Control subjects were required to have their first office encounter within one year of the matching HF case patient’s first office visit, and have at least one office encounter 30 days before or any time after the case’s HF diagnosis date to ensure similar duration of observations among cases and controls.


\section{Results on encounter sequence modeling}
\label{sec:dpm}
\textbf{Objective:} Given a sequence of visits $\xb_1, \ldots, \xb_T$, the goal of encounter sequence modeling is, for each time step $i$, to predict the codes occurring at the next visit $\xb_2, \ldots, \xb_{T+1}$. In this experiment, we focus on predicting the diagnosis codes in the encounter sequence, so we create a separate set of labels $\yb_1, \ldots, \yb_T$ that do not contain non-diagnosis codes such as medication codes or procedure codes. Therefore $\yb_i$ will contain diagnosis codes from the next visit $\xb_{i+1}$.

\textbf{Dataset:} We divide the entire dataset described in Table \ref{table:data} into 0.75:0.10:0.15 ratio, respectively for training set, validation set, and test set. 

\textbf{Baseline:} We use the same baseline models we used for HF prediction. However, since we are predicting 283 binary labels now, we replace the logistic regression function with the Softmax function. The drop-out and $L_2$ regularization policies remain the same.

For LR and MLP, at each step $i$, we aggregate maximum ten past input vectors\footnote{We also tried aggregating all past input vectors $\xb_1, \ldots, \xb_{i}$, but the performance was slightly worse than using just ten.} $\xb_{i-9}, \ldots, \xb_{i}$ to create a pseudo-context vector $\widehat{\cbb}_{i}$. LR applies the Softmax function on top of $\widehat{\cbb}_{i}$. MLP places a hidden layer on top of $\widehat{\cbb}_{i}$ then applies the Softmax function. 


\textbf{Evaluation metric:} We use the negative log likelihood Eq~\eqref{eq:cross_entropy} on the test set to evaluate the model performance. We also use Recall$\bm{@k}$ as an additional metric to measure the prediction accuracy.
\begin{itemize}[leftmargin=5.5mm]
\item
\textbf{Recall$\bm{@k}$:}
Given a sequence of visits $\xb_1, \ldots, \xb_T$, we evaluate the model performance based on how accurately it can predict the diagnosis codes $\yb_1, \ldots, \yb_T$. We use the average Recall$\bm{@k}$, which is expressed as below,
\begin{equation*}
\frac{1}{N} \sum_{n=1}^{N} \frac{1}{T^{(n)}} \sum_{i=1}^{T^{(n)}} \text{Recall}\bm{@k}(\widehat{\yb}_i),  \quad \text{\textit{where} } \quad \text{Recall}\bm{@k}(\widehat{\yb}_{i}) = \frac{| \mathrm{argsort}(\widehat{\yb}_{i})[:k] \cap nonzero(\yb_i)|}{|nonzero(\yb_i)|}
\end{equation*}
where $argsort$ returns a list of indices that will decrementally sort a given vector and $nonzero$ returns a list of indices of the coordinates with non-zero values. We use Recall$\bm{@k}$ because of its similar nature to the way a human physician performs the differential diagnostic procedure, which is to generate a list of most likely diseases for an undiagnosed patient, then perform medical practice until the true disease, or diseases are determined.
\end{itemize}

\textbf{Prediction accuracy:} Table \ref{tab:dpm} displays the prediction performance of \alg and the baselines. We use $k=5,10$ for Recall$\bm{@k}$ to allow a reasonable number of prediction trials, as well as cover complex patients who often receive multiple diagnosis codes at a single visit. 

RNN shows the best prediction accuracy for encounter diagnosis prediction. However, considering the purpose of encounter diagnosis prediction, which is to assist doctors to provide quality care for the patient, black-box behavior of RNN makes it unattractive as a clinical tool. On the other hand, \alg performs as well as other attention models, only slightly inferior to RNN, provides full interpretation of its prediction behavior, making it a feasible solution for clinical applications. 

The interesting finding in Table \ref{tab:dpm} is that MLP is able to perform as accurately as RNN+$\alpha_M$ in terms of Recall\textbf{@}10. Considering the fact that MLP uses aggregated information of past ten visits, we can assume that encounter diagnosis prediction depends more on the frequency of disease occurrences rather than the order in which they occurred. This is quite different from the HF prediction task, where stationary models (LF, MLP) performed significantly worse than sequential models. 

\begin{table}
\centering 
\caption{Encounter diagnosis prediction performance of \alg and the baselines}
    \begin{tabular}{l|c|c|c}
      {\footnotesize Model} & \begin{tabular}{@{}c@{}}{\footnotesize Negative} \\ {\footnotesize Likelihood}\end{tabular} & {\footnotesize Recall@5} & {\footnotesize Recall@10} \\
      \hline \hline
      LR & 0.0288 & 43.15 & 55.84 \\
      MLP & 0.0267 & 50.72 & 65.02 \\
      \hline
      RNN & \textbf{0.0258} & \textbf{55.18} & \textbf{69.09} \\
      \hline
      RNN+$\alpha_M$ & 0.0262 & 52.15 & 65.81 \\

      RNN+$\alpha_R$ & 0.0259 & 53.89 & 67.45 \\
      \hline
      \alg & 0.0259 & 54.25 & 67.74 \\
            \hline
    \end{tabular}
    \label{tab:dpm}
\end{table}

\section{Illustration and comparison of the baselines}
\label{sec:baseFigs}
\begin{figure}[t]
    \centering
    \begin{subfigure}[b]{0.08\textwidth}
        \includegraphics[scale=0.5]{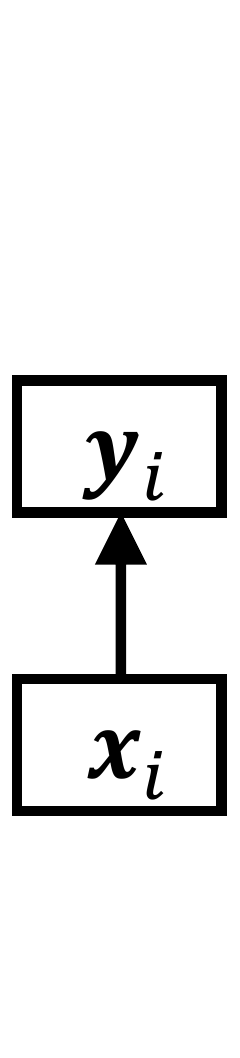}
        \caption{}
        \label{fig:LR}
    \end{subfigure}
    ~ 
    \begin{subfigure}[b]{0.09\textwidth}
        \includegraphics[scale=0.5]{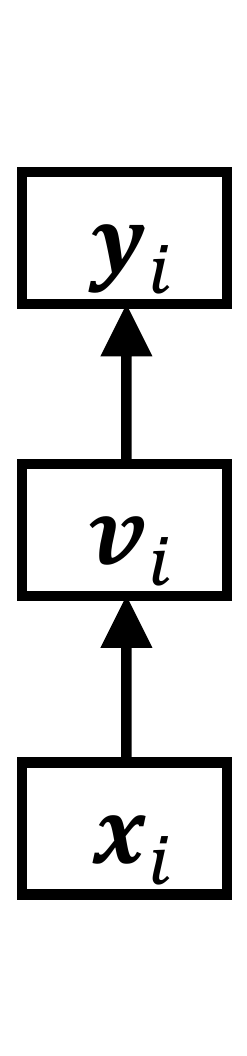}
        \caption{}
        \label{fig:mlp}
    \end{subfigure}
    ~ 
    \begin{subfigure}[b]{0.14\textwidth}
        \includegraphics[scale=0.5]{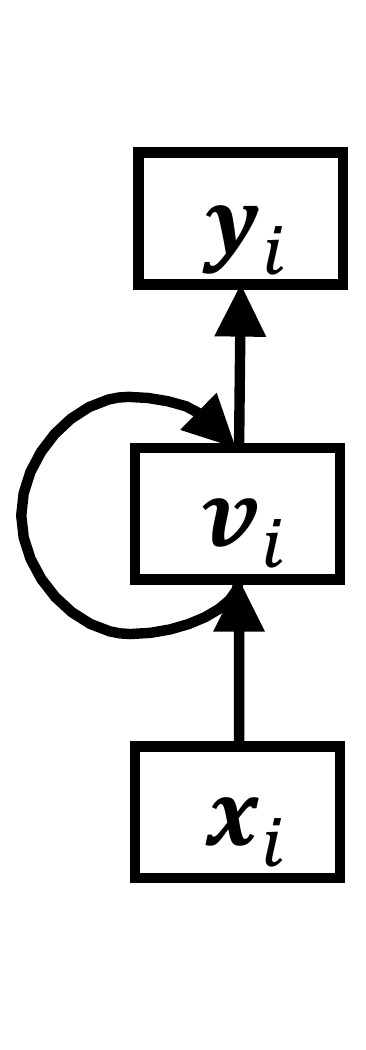}
        \caption{ }
        \label{fig:rnn}
    \end{subfigure}
     ~ 
    \begin{subfigure}[b]{0.3\textwidth}
        \includegraphics[scale=0.5]{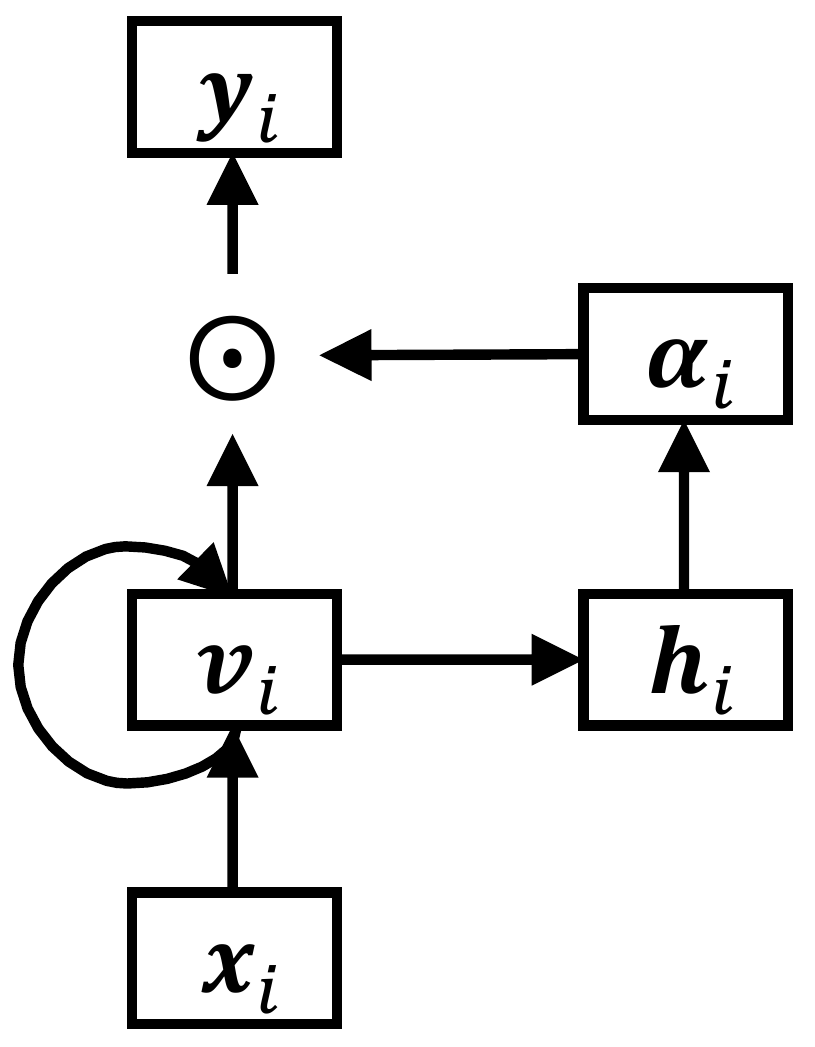}
        \caption{ }
        \label{fig:rnn-mlp}
    \end{subfigure}
     ~ 
    \begin{subfigure}[b]{0.3\textwidth}
        \includegraphics[scale=0.5]{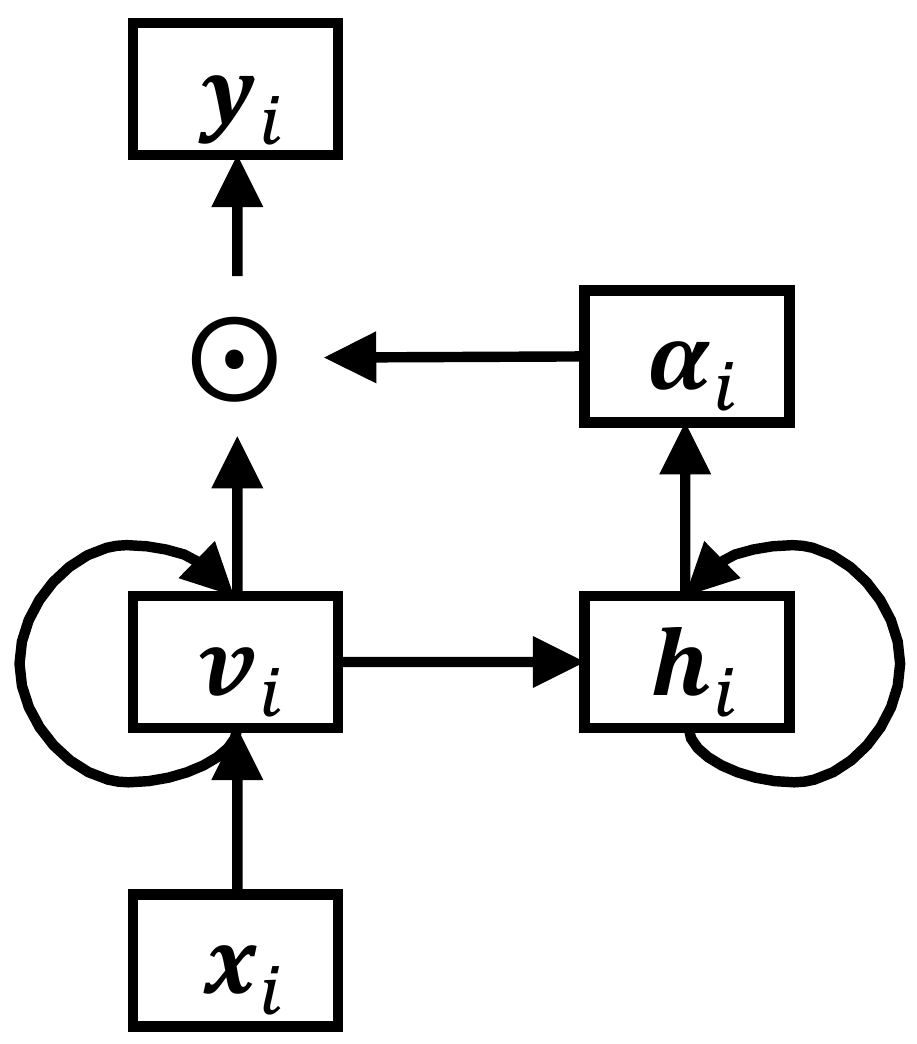}
        \caption{ }
        \label{fig:rnn-rnn}
    \end{subfigure}
    \caption{Graphical illustration of the baselines: (\subref{fig:LR}) Logistic regression (LR), (\subref{fig:mlp}) Multilayer Perceptron (MLP), (\subref{fig:rnn}) Recurrent neural network (RNN), (\subref{fig:rnn-mlp}) RNN with attention vectors generated via an MLP (RNN$+\bm{\alpha}_M$), (\subref{fig:rnn-rnn}) RNN with attention vectors generated via an RNN (RNN$+\bm{\alpha}_R$). \alg is given in Figure \ref{fig:retain}.}
    \label{fig:baselines}
\end{figure}

Figure \ref{fig:baselines} illustrates the baselines used in the experiments and shows the relationship among them.

\end{document}